\documentclass[]{spie}  %>>> use for US letter paper
\usepackage{siunitx}
\usepackage{amsmath,amsfonts,amssymb}
\usepackage{gensymb}
\newcommand{\Loss}{\mathcal{L}}
\usepackage{graphicx}
\usepackage{float}
\usepackage{svg}
\usepackage[colorlinks=true, allcolors=blue]{hyperref}

\title{RibPull: Implicit Occupancy Fields and Medial Axis Extraction for CT Ribcage Scans}

\author[a]{Emmanouil Nikolakakis}
\author[b]{Amine Ouasfi}
\author[c]{Julie Digne}
\author[d]{Razvan Marinescu}
\affil[a]{Electrical and Computer Engineering Department, University of California, Santa Cruz}
\affil[b]{Inria, Univ. Rennes, CNRS, IRISA, M2S, Rennes}
\affil[c]{LIRIS - CNRS - Université Claude Bernard Lyon 1}
\affil[d]{Computer Science Engineering Department, University of California, Santa Cruz}

\begin{document} 
\maketitle

\begin{abstract}
We present RibPull, a methodology that utilizes implicit occupancy fields to bridge computational geometry and medical imaging. Implicit 3D representations use continuous functions that handle sparse and noisy data more effectively than discrete methods. While voxel grids are standard for medical imaging, they suffer from resolution limitations, topological information loss, and inefficient handling of sparsity. Coordinate functions preserve complex geometrical information and represent a better solution for sparse data representation, while allowing for further morphological operations. Implicit scene representations enable neural networks to encode entire 3D scenes within their weights. The result is a continuous function that can implicitly compesate for sparse signals and infer further information about the 3D scene by passing any combination of 3D coordinates as input to the model. In this work, we use neural occupancy fields that predict whether a 3D point lies inside or outside an object to represent CT-scanned ribcages. We also apply a Laplacian-based contraction to extract the medial axis of the ribcage, thus demonstrating a geometrical operation that benefits greatly from continuous coordinate-based 3D scene representations versus voxel-based representations. We evaluate our methodology on 20 medical scans from the RibSeg dataset, which is itself an extension of the RibFrac dataset. We will release our code upon publication.
\end{abstract}


\section{Description of Purpose}
\label{sec:intro}  % \label{} allows reference to this section

\begin{figure}[H]
    \begin{center}
    \includegraphics[width=0.94\textwidth]{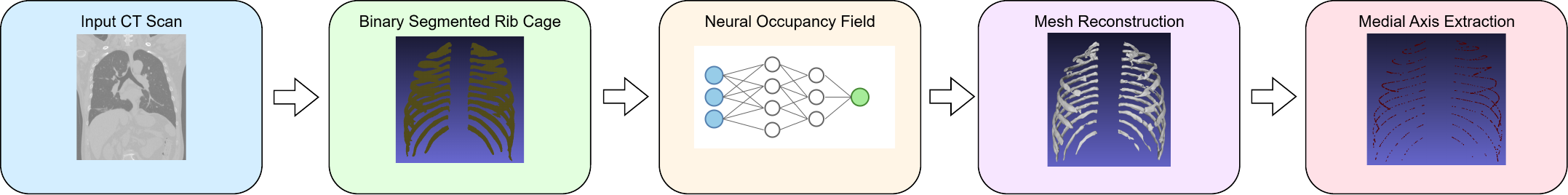}
    \end{center}
    \caption{\textbf{RibPull methodology overview.} From left to right: CT scan input, binary ribcage segmentation, neural SDF training, isosurface extraction, and medial axis extraction for morphological analysis.}\label{fig:overview}
\end{figure}

Medical imaging relies on discrete voxel representations from a variety of different scanners, such as computed tomography (CT) or positron emission tomography (PET), to provide anatomical and functional information. Although this standardized format enables straightforward segmentation and classification tasks, voxel grids impose fundamental limitations on geometric analysis due to their discrete nature and fixed resolution. These limitations prevent the extraction of clinically valuable morphological properties, such as precise surface curvature, medial axes, and smooth deformation fields, which could enhance diagnostic capabilities and provide a real-time, accurate, and interpretable visualization of the ground truth. Morphological operations like skeletonization require smooth interpolation between data points to accurately capture the underlying geometry, which is challenging with discrete voxel grids that introduce artifacts at grid boundaries. While there are a few methodologies that leverage skeletonization \cite{sato2000teasar, fuzzyskeleton}, the Laplace-Beltrami operator \cite{laplacebeltramineocortex, laplacebeltramicorpus}, and other shape analysis techniques, the medical field would strongly benefit from continuous coordinate-based representations such as Signed Distance Fields (SDF) that enable seamless interpolation at arbitrary resolutions. \cite{lesa} \\
Recent advances in neural 3D scene representation have demonstrated the potential of storing images as continuous implicit \cite{nerf} or explicit \cite{gaussiansplatting} functions. When using neural fields, the goal is to overfit multilayer perceptrons to encode entire 3D scenes within their weights. These continuous representations naturally support interpolation by providing smooth function values at any 3D coordinate, making them ideal for geometric operations that require precise spatial derivatives and smooth transitions. These approaches offer reduced memory requirements and enable novel querying capabilities such as novel view synthesis and point interpolation. However, existing applications of neural 3D representations in medical imaging \cite{nikolakakis2024gaspct, corona2022mednerf} focus primarily on angular acquisitions for sparse view acquisitions rather than geometric analysis applications. \\
This work presents RibPull, a methodology that generates implicit occupancy field representations, which are then converted to SDF representations of segmented ribcages from CT scans. CT was chosen for its ability to provide detailed anatomical and structural information of bone tissue through measurement of attenuation coefficients. Our approach first learns binary occupancy probabilities from sparse point clouds, then converts these to signed distance fields for geometric analysis. Unlike previous approaches, our method specifically targets geometric analysis applications, enabling resolution-independent queries, direct medial axis extraction, and smooth morphological operations previously impossible with discrete voxel data. The continuous nature of our representation allows for precise interpolation during skeletonization, avoiding the staircase artifacts and topological inconsistencies common in voxel-based approaches. We validate our approach using 20 segmented ribcages from the RibFrac challenge dataset \cite{jin2020deep}. \\
Our results demonstrate that SDF-based representations advance morphological analysis capabilities in medical imaging while maintaining computational efficiency. We are also able to create a representation that reduces the memory storage of the 3D scene by $\sim$ 57\%, from an average of 4.2 MB of the input point cloud training data to 1.8 MB used by the model's weights. This work bridges computational geometry and medical imaging, opening new possibilities for fracture analysis, surgical planning, and biomechanical modeling of anatomical structures. An overview of our methodology can be seen in Figure \ref{fig:overview}.

\section{Method}

\subsection{RibSeg}

RibSeg \cite{yang2021ribseg} is a ribcage segmentation model that takes a volumetric CT scan as input and converts it to a point cloud work such as PointNet++ \cite{pointent++}. More specifically, the segmentation preserves the floating ribs while removing the background structures such as the clavicle, scapula, or sternum. To train the model, RibSeg leverages the ground truth of the segmented ribcages, which is created using radiologists to annotate the ribcage manually.

\subsection{SparseOcc}
Given a point cloud $\mathcal{P}$ obtained from a volumetric CT scan using RibSeg, we propose to learn a neural implicit representation of the corresponding shape that can be used to perform geometrical operations on medical scans. In this context, the signed distance function (SDF) is the most popular representation. For an object $\Omega$, the signed distance function represents the shortest distance from any point to the object's surface, with negative values inside the object, zero on the boundary, and positive values outside. The key property of SDFs is that $|\nabla f_{\Omega}(\mathbf{x})| = 1$ almost everywhere, known as the eikonal equation.

However, without dense ground-truth supervision, learning SDFs from point clouds remains a challenging problem. To stabilize the training, different smoothness priors leveraging, for example, the gradient \cite{igr} or the Laplacian \cite{ben2022digs} of the field, have been introduced. In contrast, Ouasfi et al. \cite{ouasfi2024unsupervised} suggested the use of binary occupancy fields, which are simpler to learn as a binary signal compared to the continuous SDF field that needs to satisfy additional constraints such as the eikonal equation \cite{igr}. The proposed method defines the margin uncertainty $U_{\theta}(\mathbf{x})$ of a neural occupancy function as the difference between the probability of the query point $\mathbf{x}$ in $\mathbb{R}^3$ being outside and its probability of being inside the shape according to this neural occupancy function. It shows that sampling from the zeros of this function corresponds to sampling from the surface of the shape. Subsequently, the neural occupancy function can be supervised by minimizing the distance between the sampled points and the input point cloud $\mathcal{P}$. Samples from the surface of the occupancy function are obtained using a Newton-Raphson iteration on the margin uncertainty function $U_{\theta}(\mathbf{x})$. The corresponding loss is as follows:

\begin{equation}
\mathcal{L}_{samp} (\theta, Q) = \underset{p,q\sim Q}{\mathbb{E}} ||q - U\theta(q) \cdot \frac{\nabla U_\theta(q)}{||\nabla U_\theta(q)||_2} - p||_2^2,
\end{equation}

where $Q$ consists of pairs of query points $q$ sampled around the input point cloud and their closest input point in $\mathcal{P}$. To stabilize the training, an entropy-based regularization loss $\lambda \Loss_{entr}$ is introduced. It guides the occupancy field toward a minimal entropy almost everywhere in space and maximizes its entropy at the input point cloud. The final loss is the following:

%As no ground-truth occupancy information is available during the optimization, the proposed neural occupancy %function is trained 

\begin{equation}
   \min_{\theta} \Loss_{samp}(\theta, Q) + \lambda \Loss_{entr}(\theta, \Omega, \mathcal{P})
\end{equation}

\subsection{Laplacian-Based Contraction}

A direct benefit of continuous coordinate-based representations is that they allow for geometrical operations on medical scans. We demonstrate this capability by performing point cloud skeletonization of the ISR-encoded ribcages.
Among a variety of applicable skeletonization methodologies \cite{sato2000teasar, huang2013l1, clemot2023neural}. We choose a Laplacian-based contraction\cite{laplacecontraction} due to its robustness, simplicity, and interpretability. We also observe that it performs particularly well with sparse and noisy data. This algorithm is based on contracting the shape while preserving the geometry as dictated by the following linear system of equations:

\begin{equation}
\begin{bmatrix}
W_L L \\
W_H
\end{bmatrix}
\mathbf{P}' = 
\begin{bmatrix}
\mathbf{0} \\
W_H \mathbf{P}
\end{bmatrix}
\label{eq:laplacian_contraction}
\end{equation}

\begin{equation}
E(\mathbf{P}') = \|W_L L \mathbf{P}'\|_F^2 + \sum_{i=1}^{n} W_{H,i}^2 \|\mathbf{p}'_i - \mathbf{p}_i\|_2^2
\label{eq:contraction_energy}
\end{equation}

where $L \in \mathbb{R}^{n \times n}$ is the Laplacian matrix with cotangent weights, $\mathbf{P}' \in \mathbb{R}^{n \times 3}$ represents the contracted point cloud, $W_L, W_H \in \mathbb{R}^{n \times n}$ are diagonal weight matrices balancing contraction and attraction forces, and $\|\cdot\|_F$ denotes the Frobenius norm and $\|\cdot\|_2$ the Euclidean norm.

\section{Results}

\subsection{Dataset}

The datasets we use for this work are directly provided by RibSeg. RibSeg leverages CT scans provided by ribfrac \cite{jin2020deep}, which consists of a large collection of 3D DICOM CT chest scans. To validate the segmentation's accuracy against ground truth images, RibSeg used point clouds that were manually annotated by radiologists. These annotated point clouds are used for our experiments due to their accuracy and structural consistency.

\subsection{Results}

\begin{figure}[H]
    \begin{center}
        \includegraphics[width=.7\textwidth]{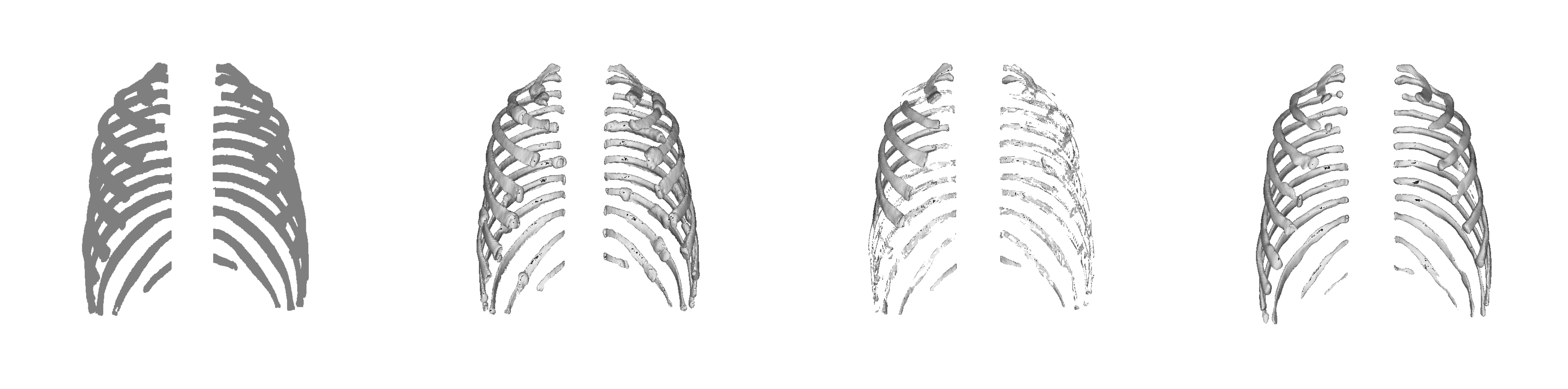}
       
        \makebox[.16\textwidth]{Ground Truth} \makebox[.16\textwidth]{DiGS} \makebox[.16\textwidth]{N-Pull} \makebox[.16\textwidth]{SparseOcc}
    \end{center}
    \caption{Ribcage reconstruction comparison. Ground truth segmented ribcage from RibSeg dataset (left), followed by reconstructions using DiGs, N-Pull, and SparseOCC methods, respectively. The progression shows surface reconstruction to skeleton extraction using occupancy field-based approaches.\label{fig:isrcomp}}
\end{figure}

\begin{figure}[H]
    \begin{center}
        \includegraphics[trim=0mm 14mm 0mm 12mm, clip, width=.7\textwidth]{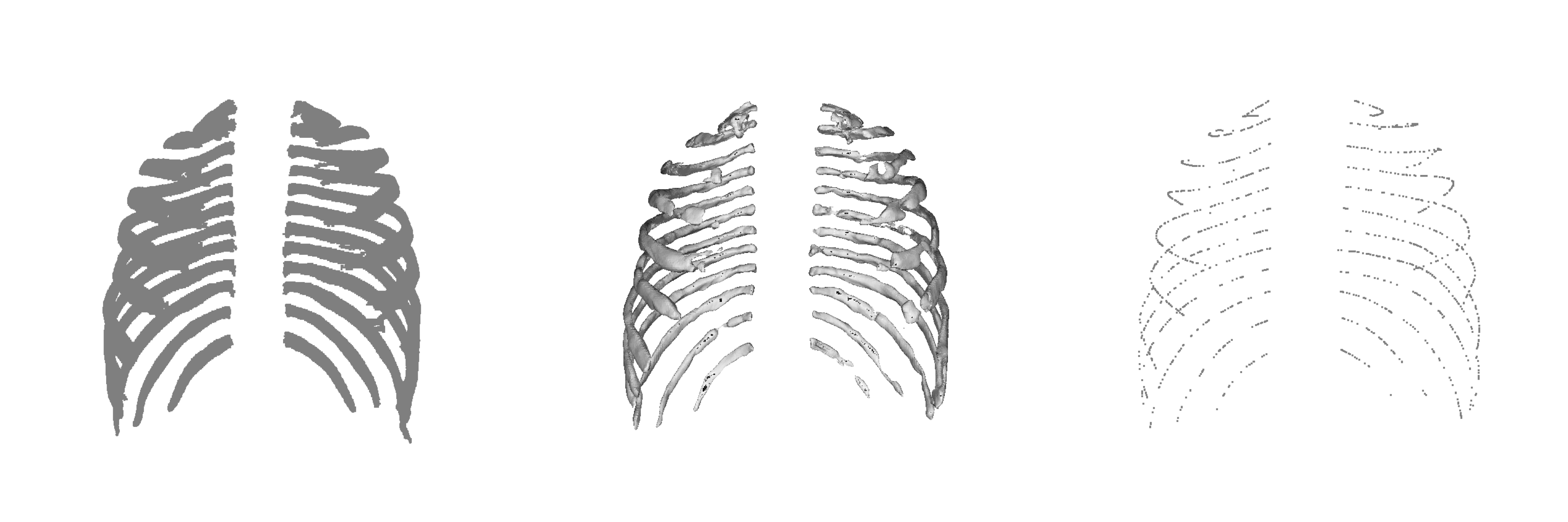}
       
        \makebox[.22\textwidth]{Ground Truth} \makebox[.26\textwidth]{Reconstruction} \makebox[.22\textwidth]{Skeletonization}
    \end{center}
    \caption{On the left, ground truth segmented ribcage using RibSeg on a dataset provided by the ribfrac challenge. In the middle is a surface reconstruction of the implicit SDF using the Marching Cubes algorithm, with the threshold being set to the median uncertainty of the Occupancy Field. On the right, the final skeleton using Laplacian-based Contraction for medial axis extraction}\label{fig:pipeline}
\end{figure}

We compare our methodology against DiGS \cite{ben2022digs} and Neurall-Pull \cite{neuralpull}. Our reconstruction results can be observed in Figure \ref{fig:isrcomp}, while the skeletonization output is demonstrated in Figure \ref{fig:pipeline}. The metrics in table \ref{tab:metrics} show that RibPull performs the best in Chamfer-L1 and Chamfer-L2, while NeuralPull is superior in Hausdorff Distance.

\begin{table}[h]
\centering
\caption{Performance of the Uncertainty-based Occupancy Field extraction. We compare the distances between the ground truth and the reconstructed point clouds after sampling 100000 points with normalization applied.}
\begin{tabular}{||l|c|c|c||}
\hline
Metric & DiGS & NeuralPull & SparseOcc \\
\hline
Chamfer-L1 ↓ & 8.04 ± 10.1 & 5.81 ± 0.38 & \textbf{1.55 ± 0.24} \\
 \hline
Chamfer-L2 ↓ & 5.36 ± 6.71 & 3.90 ± 0.26 & \textbf{1.04 ± 0.16} \\
 \hline
Hausdorff ↓ & 36.8 ± 23.0 & \textbf{8.25 ± 3.22} & 12.6 ± 4.39 \\
\hline
\end{tabular}
\label{tab:metrics}
\end{table}

\section{New or Breakthrough Work to be Presented}

We introduce RibPull, the first implicit surface reconstruction (ISR) methodology for medical imaging that leverages occupancy fields to reconstruct complex anatomical structures from CT scans. Our approach employs a neural pull-based optimization framework with minimax entropy regularization to handle the challenging thin structures characteristic of skeletal anatomy. The methodology is specifically designed to handle the noise and sparsity inherent in medical imaging data. We demonstrate the effectiveness of our approach on computed tomography ribcage reconstructions. This work establishes the foundation for future extensions to compressed sensing scenarios and broader CT anatomical structure reconstruction. constraints.

\section{Conclusions}
Anatomy preservation is critical for CT scene representations. Following the training of the neural occupancy fields, we observe that SparseOcc outperforms other methodologies by accurately preserving the topological and geometrical structure of the object. This is expected as SparseOcc is specifically designed for sparse and noisy point clouds, while dense point cloud models typically collapse and are less robust under such conditions. \\
Our methodology provides a simple, minimal, and interpretable visualization of ribcage anatomy that enables fracture detection and scoliosis assessment, while serving as a potential tool for surgical planning. The Laplacian-based contraction performs well for medial axis extraction, demonstrating the advantages of continuous representations for geometric operations that would be challenging with traditional voxel-based approaches. \\
However, several limitations remain to be addressed. The reconstruction accuracy does not yet reach desired levels, and some information loss occurs during training. Additionally, our evaluation was performed on datasets that have undergone manual curation by radiologists, which may not reflect real-world clinical scenarios with inherent noise and artifacts. Future work will focus on improving 3D scene encoding and skeletonization performance and testing our methodology on noisy CT image scans. Finally, we intend to investigate the preservation of geometric properties such as surface normals and curvature under extreme compressed sensing constraints, which could further expand the clinical applicability of our approach.

This work is not being, or has been, submitted for publication or presentation elsewhere.

\pagebreak

\section{Acknowledgements}

We gratefully acknowledge the authors of RibSeg \cite{yang2021ribseg, yang2023ribseg} for making their large-scale benchmark dataset publicly available. This work utilized the rib segmentation annotations from RibSeg, which extends 660 CT scans from the RibFrac dataset with detailed rib labeling and anatomical centerlines. We thank the creators of both the RibFrac and RibSeg datasets for their valuable contributions to the medical imaging research community. We are also planning to make our code public following acceptance of our work.

% References
\bibliography{report} % bibliography data in report.bib

\begin{thebibliography}{10}

\bibitem{sato2000teasar}
Sato, M., Bitter, I., Bender, M.~A., Kaufman, A.~E., and Nakajima, M., ``Teasar: tree-structure extraction algorithm for accurate and robust skeletons,'' in [{\em Proceedings the Eighth Pacific Conference on Computer Graphics and Applications}{\nolinebreak\hspace{0.1em}]},   281--449, IEEE (2000).

\bibitem{fuzzyskeleton}
Jin, D. and Saha, P.~K., ``A new fuzzy skeletonization algorithm and its applications to medical imaging,'' in [{\em International Conference on Image Analysis and Processing}{\nolinebreak\hspace{0.1em}]},   662--671, Springer (2013).

\bibitem{laplacebeltramineocortex}
Qiu, A., Bitouk, D., and Miller, M.~I., ``Smooth functional and structural maps on the neocortex via orthonormal bases of the laplace-beltrami operator,'' {\em IEEE Transactions on Medical Imaging}~{\bf 25}(10),  1296--1306 (2006).

\bibitem{laplacebeltramicorpus}
Shi, Y., Lai, R., Krishna, S., Sicotte, N., Dinov, I., and Toga, A.~W., ``Anisotropic laplace-beltrami eigenmaps: bridging reeb graphs and skeletons,'' in [{\em 2008 IEEE computer society conference on computer vision and pattern recognition workshops}{\nolinebreak\hspace{0.1em}]},   1--7, IEEE (2008).

\bibitem{lesa}
Zhang, Z., Wu, Y., Xiong, D., Ibrahim, J.~G., Srivastava, A., and Zhu, H., ``Lesa: Longitudinal elastic shape analysis of brain subcortical structures,'' {\em Journal of the American Statistical Association}~{\bf 118}(541),  3--17 (2023).

\bibitem{nerf}
Mildenhall, B., Srinivasan, P.~P., Tancik, M., Barron, J.~T., Ramamoorthi, R., and Ng, R., ``Nerf: Representing scenes as neural radiance fields for view synthesis,'' {\em Communications of the ACM}~{\bf 65}(1),  99--106 (2021).

\bibitem{gaussiansplatting}
Kerbl, B., Kopanas, G., Leimk{\"u}hler, T., and Drettakis, G., ``3d gaussian splatting for real-time radiance field rendering,'' {\em ACM Transactions on Graphics}~{\bf 42}(4) (2023).

\bibitem{nikolakakis2024gaspct}
Nikolakakis, E., Gupta, U., Vengosh, J., Bui, J., and Marinescu, R., ``Gaspct: Gaussian splatting for novel brain cbct projection view synthesis,'' in [{\em Medical Imaging 2025: Image Processing}{\nolinebreak\hspace{0.1em}]},   {\bf 13406},  318--325, SPIE (2025).

\bibitem{corona2022mednerf}
Corona-Figueroa, A., Frawley, J., Bond-Taylor, S., Bethapudi, S., Shum, H.~P., and Willcocks, C.~G., ``Mednerf: Medical neural radiance fields for reconstructing 3d-aware ct-projections from a single x-ray,'' in [{\em 2022 44th Annual International Conference of the IEEE Engineering in Medicine \& Biology Society (EMBC)}{\nolinebreak\hspace{0.1em}]},   3843--3848, IEEE (2022).

\bibitem{jin2020deep}
Jin, L., Yang, J., Kuang, K., Ni, B., Gao, Y., Sun, Y., Gao, P., Ma, W., Tan, M., Kang, H., et~al., ``Deep-learning-assisted detection and segmentation of rib fractures from ct scans: Development and validation of fracnet,'' {\em EBioMedicine}~{\bf 62} (2020).

\bibitem{yang2021ribseg}
Yang, J., Gu, S., Wei, D., Pfister, H., and Ni, B., ``Ribseg dataset and strong point cloud baselines for rib segmentation from ct scans,'' in [{\em Medical Image Computing and Computer Assisted Intervention--MICCAI 2021: 24th International Conference, Strasbourg, France, September 27--October 1, 2021, Proceedings, Part I 24}{\nolinebreak\hspace{0.1em}]},   611--621, Springer (2021).

\bibitem{pointent++}
Qi, C.~R., Yi, L., Su, H., and Guibas, L.~J., ``Pointnet++: Deep hierarchical feature learning on point sets in a metric space,'' {\em Advances in neural information processing systems}~{\bf 30} (2017).

\bibitem{igr}
Gropp, A., Yariv, L., Haim, N., Atzmon, M., and Lipman, Y., ``Implicit geometric regularization for learning shapes,'' (2020).

\bibitem{ben2022digs}
Ben-Shabat, Y., Koneputugodage, C.~H., and Gould, S., ``Digs : Divergence guided shape implicit neural representation for unoriented point clouds,'' (2023).

\bibitem{ouasfi2024unsupervised}
Ouasfi, A. and Boukhayma, A., ``Unsupervised occupancy learning from sparse point cloud,'' in [{\em Proceedings of the IEEE/CVF Conference on Computer Vision and Pattern Recognition}{\nolinebreak\hspace{0.1em}]},   21729--21739 (2024).

\bibitem{huang2013l1}
Huang, H., Wu, S., Cohen-Or, D., Gong, M., Zhang, H., Li, G., and Chen, B., ``L1-medial skeleton of point cloud.,'' {\em ACM Trans. Graph.}~{\bf 32}(4),  65--1 (2013).

\bibitem{clemot2023neural}
Cl{\'e}mot, M. and Digne, J., ``Neural skeleton: Implicit neural representation away from the surface,'' {\em Computers \& Graphics}~{\bf 114},  368--378 (2023).

\bibitem{laplacecontraction}
Cao, J., Tagliasacchi, A., Olson, M., Zhang, H., and Su, Z., ``Point cloud skeletons via laplacian based contraction,'' in [{\em 2010 Shape Modeling International Conference}{\nolinebreak\hspace{0.1em}]},   187--197, IEEE (2010).

\bibitem{neuralpull}
Ma, B., Han, Z., Liu, Y.-S., and Zwicker, M., ``Neural-pull: Learning signed distance functions from point clouds by learning to pull space onto surfaces,'' {\em arXiv preprint arXiv:2011.13495}  (2020).

\bibitem{yang2023ribseg}
Jin, L., Gu, S., Wei, D., Adhinarta, J.~K., Kuang, K., Zhang, Y.~J., Pfister, H., Ni, B., Yang, J., and Li, M., ``Ribseg v2: A large-scale benchmark for rib labeling and anatomical centerline extraction,'' {\em IEEE Transactions on Medical Imaging}~{\bf 43}(1),  570--581 (2023).

\end{thebibliography}
\bibliographystyle{spiebib} % makes bibtex use spiebib.bst

\end{document}